\documentclass{article} 
\usepackage{iclr2020_conference,times}

\usepackage[utf8]{inputenc}

\usepackage{hyperref}
\usepackage{url}

\usepackage{xcolor}

\usepackage{amsmath,amsfonts,amssymb,amsthm}
\usepackage{booktabs}
\usepackage{marvosym}

\usepackage{gensymb}
\usepackage{subcaption}

\usepackage[ruled, vlined, linesnumbered]{algorithm2e}
\SetKwInput{KwData}{Inputs}
\SetKwInput{KwResult}{Output}

\usepackage{cleveref}
\crefformat{equation}{(#2#1#3)}
\crefformat{figure}{Figure #2#1#3}

\usepackage[super]{nth}
\usepackage{mathtools}
\usepackage{amsmath,amsfonts,amssymb,amsthm, bm, bbm}
\usepackage{textcomp}
\usepackage{nicefrac}       
\usepackage{placeins}  

\newcommand{\bff}{\mathbf}
\DeclarePairedDelimiterX{\norm}[1]{\lVert}{\rVert}{#1}

\newcommand{\dset}{{\mathfrak D}}

\DeclareMathOperator*{\argmax}{arg\,max}

\newcommand{\balpha}{\bm{\alpha}}
\newcommand{\bbeta}{\boldsymbol{\beta}}

\newcommand{\btheta}{\boldsymbol{\theta}}

\newcommand{\RN}[1]{%
  \textup{\uppercase\expandafter{\romannumeral#1}}%
}
\DeclareMathOperator{\EX}{\mathbb{E}}

\usepackage{tikz}
\usetikzlibrary{arrows}
\usetikzlibrary{bayesnet}
\usetikzlibrary{chains, positioning, decorations.pathreplacing}
    
\tikzset{nicecolor/.style={draw=#1!50!black, top color=#1!70!black!20!white, bottom color=#1!70!black!20!white}}
    
\tikzset{opnode/.style={circle, minimum width=3ex}}
\tikzset{plus/.style ={opnode, nicecolor=blue, font={$+$}}}
\tikzset{times/.style={opnode, nicecolor=blue, font={$\times$}}}

\tikzset{block/.style={rectangle, text width=5ex, align=center, minimum height=0.7cm, nicecolor=red, font=\scshape, rounded corners}}

\tikzset{line/.style={thick, draw=gray!80!black}}
\tikzset{arrow/.style={-stealth,line}}

\title{Variational Depth Search in ResNets} 

\author{Javier Antorán\thanks{equal contribution} \\ 
University of Cambridge \\
\texttt{ja666@cam.ac.uk} \\
\And
James Urquhart Allingham\footnotemark[\value{footnote}] \\ 
University of Cambridge \\
\texttt{jua23@cam.ac.uk} \\
\And
José Miguel Hernández-Lobato \\ 
University of Cambridge \\
Microsoft Research \\
The Alan Turing Institute \\
\texttt{jmh233@cam.ac.uk}
}

\iclrfinalcopy 
\begin{document}

\maketitle

\begin{abstract}
One-shot neural architecture search allows joint learning of weights and network architecture, reducing computational cost. We limit our search space to the depth of residual networks and formulate an analytically tractable variational objective that allows for obtaining an unbiased approximate posterior over depths in one-shot. We propose a heuristic to prune our networks based on this distribution. We compare our proposed method against manual search over network depths on the MNIST, Fashion-MNIST, SVHN datasets. We find that pruned networks do not incur a loss in predictive performance, obtaining accuracies competitive with unpruned networks. Marginalising over depth allows us to obtain better-calibrated test-time uncertainty estimates than regular networks, in a single forward pass.

\end{abstract}
\section{Introduction and Related Work}\label{sec:intro}

One-shot Neural Architecture Search (NAS) is a promising approach to NAS that uses weight-sharing to significantly reduce the computational cost of exploring the architecture search space. This makes NAS more accessible to researchers and practitioners without large computational budgets. 
In this work, we describe a computationally cheap, gradient-based, one-shot NAS method that uses Variational Inference~(VI) to learn distributions over the depth of residual networks (ResNets). Our approach inherits advantages from Bayesian neural networks such as capturing model uncertainty and robustness to over-fitting~\citep{hernndezlobato2015probabilistic,Gal2016Uncertainty}.

Perhaps the most well known gradient-based one-shot NAS approach is DARTS~\citep{darts}. It uses a continuous relaxation of the search space to learn the structure of \emph{cells} within a larger, fixed, computational graph. 
Each edge in the graph of a cell represents a mixture of possible operations. Mixture weights are optimised with respect to the validation set. SNAS~\citep{xie2018snas}, ProxylessNAS~\citep{cai2018proxylessnas} and BayesNAS~\citep{zhou2019bayesnas} take similar approaches, varying the distributions over cell operations and optimisation procedures. 
\citet{shin2018differentiable} use gradients to jointly optimise weights and hyper-parameters for network layers.
\citet{maskconnect} jointly optimise graph connectivity and weights using binary variables and a modified back-propagation algorithm. In contrast, we restrict our search to network depth. 
We jointly learn model weights and a distribution over depths, as opposed to a point estimate, using only the train set.

More closely related to this work is that of \citet{dikov2019bayesian}, who learn both the depth and width of a ResNet using VI. They obtain biased estimates of the Evidence Lower BOund's (ELBO's) gradients with respect to model architecture by leveraging continuous relaxations of discrete probability distributions. 
\citet{nalisnick2018dropout} interpret dropout as a \emph{structured shrinkage prior}. They use it for \emph{automatic depth determination} in ResNets, reducing the influence of, but not removing, whole residual blocks.
\cite{bender2019understanding} use \emph{path dropout}, in which whole edges of cells are dropped out at training time, to prevent co-adaptation while performing one-shot AS. Conversely, we directly model depth as a categorical variable instead of a product of Bernoullis. As a result, we are able to evaluate our ELBO exactly and efficiently; only a single forward pass is required to evaluate the likelihood rather than high-variance Monte-Carlo sampling.




Our main contributions are as follows: 1. We propose a probabilistic model, an approximate distribution, and a network architecture that, when combined, allow for exact evaluation of the ELBO with a single forward pass through the network. Network depth and weights are optimised jointly. 2. We show how our formulation learns distributions over depths that assign more mass to better performing architectures and are amenable to layer pruning. 3. We show how to obtain model uncertainty estimates from a single forward pass through our networks.

\begin{figure}
\vspace{-0.1cm}
\centering
\begin{subfigure}[b]{0.18\textwidth}
\centering
\resizebox{1\textwidth}{!}{
    \begin{tikzpicture}
                    
              \node[obs]                                 (x1) {$\mathbf{x}_{n}$};
              \node[latent, below=of x1, xshift=0cm]             (y1) {$\mathbf{y}_{n}$};
              \node[const, left=4.5ex of y1]                      (theta) {$\theta$};
              \node[latent, right=of y1, yshift=0cm, xshift=0cm]                      (d) {$d$};
              \node[const, right=of x1, yshift=-0.4cm, xshift=0.23cm]                      (beta) {$\boldsymbol{\beta}$};
              
              \edge {x1} {y1} ; %
              \edge {theta} {y1} ;
              \edge {d} {y1} ;
              \edge {beta} {d} ;
              
              \plate [inner sep=0.4cm] {GMMplate} {(y1)(x1)} {$\quad N$} ;
            \end{tikzpicture}
            }
\end{subfigure}
\quad
\begin{subfigure}[b]{0.763\textwidth}
    \centering
    \resizebox{1\textwidth}{!}{
    \begin{tikzpicture}[scale=1]
    	\node[block] (IB) {$f_{0}$};
    	\node[block, above right=7ex and 3ex of IB] (RB1) {$f_{1}$};
    	\node[block, right=12ex of RB1] (RB2) {$f_{2}$};
    	\node[block, right=12ex of RB2] (RB3) {$f_{D}$};
    	\node[block, below right=9ex and 6ex of RB3, minimum height=1cm] (OB) {$f_{D+1}$};
    
    	\node[times, below right=1ex and 0ex of RB1] (M1) {};
    	\node[times, below right=1ex and 0ex of RB2] (M2) {};
    	\node[times, below right=1ex and 0ex of RB3] (M3) {};
    	
    	\node[right=2ex of M1] (B1) {$b_1$};
    	\node[right=2ex of M2] (B2) {$b_{2}$};
    	\node[right=2ex of M3] (B3) {$b_D$};
    	
        \node[plus] at (IB -| M1) (A1) {};
        \node[plus] at (IB -| M2) (A2) {};
        \node[plus] at (IB -| M3) (A3) {};
    	
    	\node[left=2ex of IB] (X) {$\mathbf{x}$};
    	
    	\node[right=1.5ex of A2] (ddd) {...};
    	
    	\node[right=2ex of OB.north east] (YD) {$\mathbf{\hat y}_D$};
    	\node[right=2ex of OB] (YD1) {$\mathbf{\hat y}_{1}$};
    	\node[right=2ex of OB.south east] (Y1) {$\mathbf{\hat y}_0$};
    
    	\draw[arrow] (X.east) -- (IB.west);
    	\draw[arrow] (IB.east) -- (A1);
    	
    	\draw[arrow] (RB1.east) -| (M1);
    	\draw[arrow] (B1) -- (M1);
    	\draw[arrow] (M1) -- (A1);
    	\draw[arrow] ([xshift=-12.5ex] A1.east) |- (RB1.west);
    	
    	\draw[arrow] (A1) -- (A2);
    	
    	\draw[arrow] (RB2.east) -| (M2);
    	\draw[arrow] (B2) -- (M2);
    	\draw[arrow] (M2) -- (A2);
    	\draw[arrow] ([xshift=-12.5ex] A2.east) |- (RB2.west);
    	
    	\draw[line] (A2) -- (ddd);
    	\draw[arrow] (ddd) -- (A3);
    	
    	\draw[arrow] (RB3.east) -| (M3);
    	\draw[arrow] (B3) -- (M3);
    	\draw[arrow] (M3) -- (A3);
    	\draw[arrow] ([xshift=-12.5ex] A3.east) |- (RB3.west);
    	
    	\draw[arrow, dashed] (A3) -- ([yshift=-0.5ex] OB.north west);
    	\draw[arrow, dashed] ([xshift=-29ex] A3.east) |- (OB.west);
    	\draw[arrow, dashed] ([xshift=-29ex] A2.east) |- ([yshift=0.5ex] OB.south west);
    	
    	\draw[arrow, dashed] (OB.north east) -- (YD);
    	\draw[arrow, dashed] (OB.east) -- (YD1);
    	\draw[arrow, dashed] (OB.south east) -- (Y1);
         
    \end{tikzpicture}
    }
\end{subfigure}
\caption{Left: graphical model under consideration. Right: computational model. Each layer's activations are passed through the output block, producing predictions $\bff{\hat{y}}_{i}\,{=}\,\text{softmax}(f_{D+1}(\bff{a}_{i}))$. }
    \label{fig:nn_structure_graph_model}
\vspace{-0.1in}
\end{figure}
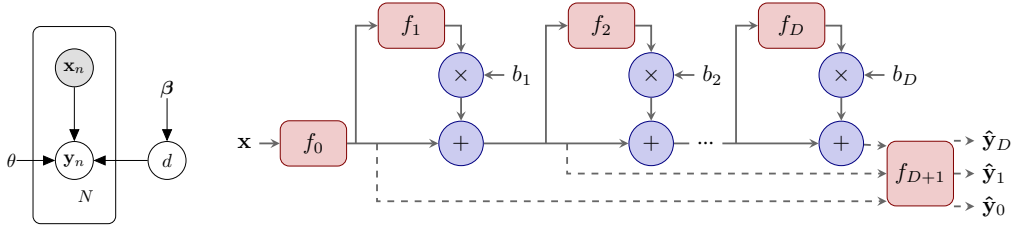

\section{Variational Inference over Architecture Space}\label{sec:proposed_method}

Consider a dataset $\dset\,{=}\,\{\bff{x}^{(n)}, \bff{y}^{(n)}\}_{n=1}^{N}$ and a neural network, parametrised by $\btheta$, formed of $D$ residual blocks $\{f_{i}(\cdot)\}_{i=1}^{D}$, an input block $f_{0}(\cdot)$, and an output block $f_{D+1}(\cdot)$. We take all layers to have a fixed width, or number of channels, $w$. We introduce a set of binary variables $\{b_{i}\}_{i=1}^{D}$ such that the activations at depth $i$, $\bff{a}_{i}$, can be obtained recursively as ${\bff{a}_{i}\,{=}\,\bff{a}_{i-1} + b_{i}\cdot f_{i}(\bff{a}_{i-1})}$. We obtain $\bff{a}_{0}$ as $f_{0}(\bff{x})$ and parameterise a distribution over targets with our model's output: $p_{\btheta}(\bff{y}|f_{D+1}(\bff{a}_{D}))$. This computational model is displayed in  \cref{fig:nn_structure_graph_model}.

Given the above formulation, we can obtain a network of depth $d$ by setting $b_{i}\,{=}\,1\,\forall\,i\leq d$. Its outputs are then given by $f_{D+1}(\bff{a}_{d})$. Deeper networks can express more complex functions but come at increased risk of overfitting and test time computational cost. We propose to manage this trade-off by placing a categorical prior distribution over depths $p_{\bbeta}(d)\,{=}\,Cat(d| \{\beta_{i}\}_{i=0}^{D})$. By selecting larger values of $\beta_{i}$ for smaller depths, we encourage simpler, computationally cheaper models. 
The posterior distribution over depths $p(d|\dset) \propto p_{\bbeta}(d)\cdot\prod_{n=1}^{N} p_{\btheta}(\bff{y}^{(n)}|\bff{x}^{(n)}, d)$ takes the form of a categorical. Unfortunately, obtaining it requires computing the likelihood of the full dataset.

We approximate the posterior distribution over depths by introducing a surrogate categorical distribution $q_{\balpha}(d) \,{=}\, Cat(d| \{\alpha_{i}\}_{i=0}^{D})$. We can optimise the variational parameters $\balpha$ and model parameters $\btheta$ simultaneously through the following objective:
\begin{gather}\label{eq:var_objective}
    \mathcal{L}(\balpha, \btheta) = \textstyle{\sum}_{n=1}^{N} \EX_{q_{\balpha}(d)}\left[ \log p_{\btheta}(\bff{y}^{(n)}|\bff{x}^{(n)}, d)\right] - \text{KL}( q_{\balpha}(d)\,\|\,p_{\bbeta}(d)).
\end{gather}
Intuitively, the first term in \cref{eq:var_objective} encourages quality of fit while the second keeps our model shallow. In \cref{app:derivation}, we link the objective in \cref{eq:var_objective} to variational EM and show it is a lower bound on $\log p(\dset)$. Because both our approximate and true posteriors are categorical, \cref{eq:var_objective} is convex w.r.t. $\balpha$. At the optima, $q_{\balpha}(d)\,{=}\,p(d|\dset)$ and the bound is tight. Thus, we are able to perform unbiased maximum likelihood estimation of network weights $\btheta$ while the depth is marginalised. Taking expectations over $q_{\balpha}(d)$ allows us to avoid calculating the exact posterior at every optimisation step.


$\EX_{q_{\balpha}(d)}[\log p_{\btheta}(\bff{y}|\bff{x}, d)]$ can be computed from the activations of every residual block. These are obtained with a single forward pass. As a result, both terms in \cref{eq:var_objective} can be evaluated exactly. This removes the need for high-variance estimators, often associated with performing VI in neural networks \citep{variational_dropout}. Using mini-batches of size $N'$, we stochastically estimate \cref{eq:var_objective} as:
\begin{gather}\label{eq:detailed_objective}
    \mathcal{L}(\balpha, \btheta) \approx \frac{N}{N'} \sum^{N'}_{n=1} \sum_{i=0}^{D} \left(\log p_{\btheta}(\bff{y}^{(n)} | f_{D+1}(\bff{a}_{i}^{(n)})) \cdot \alpha_{i} \right) - \sum_{i=0}^{D} \left( \alpha_{i} \log \frac{\alpha_{i}}{\beta_{i}} \right).
\end{gather}
After training, $q_{\balpha}(d{=}i)\,{=}\,\alpha_{i}$ represents our confidence that the number of residual blocks we should use is $i$. In low data regimes, where both terms in \cref{eq:var_objective} are of comparable scale, we choose $d_{\text{opt}}{=}\argmax_{i}\alpha_{i}$. In medium to big data regimes, where the log-likelihood dominates our objective, we find that the values of $\alpha_{i}$ flatten out after reaching an appropriate depth. Heuristically, we define $\bff{s}\,{=}\,\{i : \alpha_{i} \geq 0.95 \max_{i} \alpha_{i}\}$ and select $d_{\text{opt}}{=}\min_{i} \bff{s}$, ensuring we keep the minimum number of layers needed to explain the data well. We prune all blocks after $d_{opt}$ by setting $q_{\balpha}(d{=}d_{\text{opt}})\,{=}\, q_{\balpha}(d{\geq}d_{\text{opt}})$ and then $q_{\balpha}(d{>}d_{\text{opt}})\,{=}\,0$. Instead of also discarding the learnt probabilities over shallower networks, we incorporate them when making predictions on new data points $\bff{x}^{*}$ through marginalisation at no additional computational cost:
\begin{gather}\label{eq:predict_cutoff_depth}
    p(\bff{y}^{*} | \bff{x}^{*}) \approx \sum^{d_{\text{opt}}}_{i=0} p_{\btheta}(\bff{y}^{*} | \bff{x}^{*}, d{=}i) q_{\balpha}(d{=}i).
\end{gather}

\section{Experiments}\label{sec:experiments}

We refer to our approach as learnt depth networks (LDN). We benchmark against deterministic depth networks (DDN) for which we evaluate our search space by training networks of multiple depths. We use the same architectures and hyperparameters for LDNs and DDNs. Implementation details are given in \cref{app:implementation}. Our code can be found at: {\color{blue}\href{https://github.com/cambridge-mlg/arch_uncert}{github.com/cambridge-mlg/arch\_uncert}}.

\subsection{Spiral Classification}

We generate a 2d training set by drawing 200 samples from a 720{\degree} rotation 2-armed spiral function with additive Gaussian noise of $\sigma\,{=}\,0.15$. The test set is composed of an additional 1800 samples. We repeat experiments 6 times and report standard deviations as error bars. 
\begin{figure}[ht]
\begin{center}
\centerline{\includegraphics[width=0.95\textwidth]{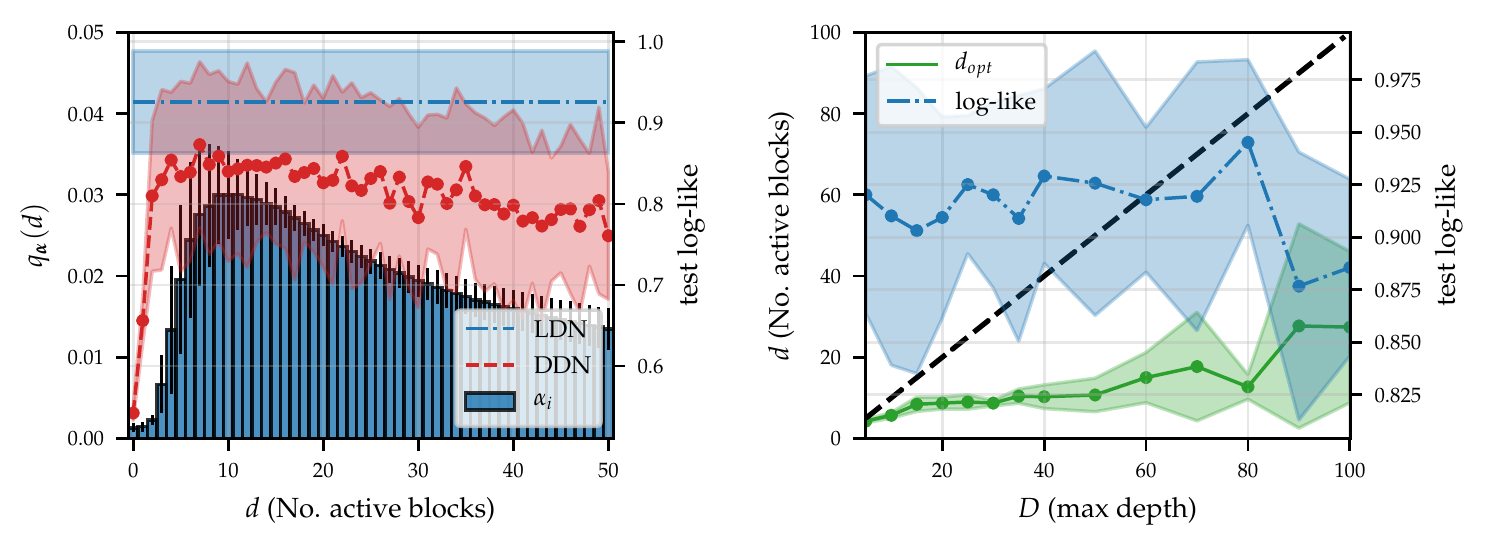}}
\vskip -0.1in
\caption{Left: posterior over depths for a LDN of $D\,{=}\,50$ trained on our spirals dataset. Test log-likelihood values obtained for DDNs at every depth are overlaid with the log-likelihood value obtained with a LDN when marginalising over $d_{opt}\,{=}\,9$ layers. Right: the LDN's chosen depth $d_{\text{opt}}{=}\argmax_{i} q_{\balpha}(d{=}i)$ and test performance remain stable as $D$ increases up until $D\,{\approx}\,50$.}
\label{fig:spiral_depth_barplot}
\end{center}
\vskip -0.2in
\end{figure}

Choosing a relatively small width $w\,{=}\,20$ to ensure the task can not be solved with a shallow model, we train LDNs with varying values of $D$ and DDNs of all depths up to $D{=}100$. \cref{fig:spiral_depth_barplot} shows how the depths to which LDNs assign larger probabilities match those at which DDNs perform best. Predictions from LDNs pruned to $d_{opt}$ layers outperform DDNs at all depths. The chosen $d_{opt}$ remains stable for increasing maximum depths up to $D\approx50$. The same is true for test performance, showing some robustness to overfitting. After this point, training starts to become unstable.

\begin{figure}[ht]
\begin{center}
\centerline{\includegraphics[width=0.8\textwidth]{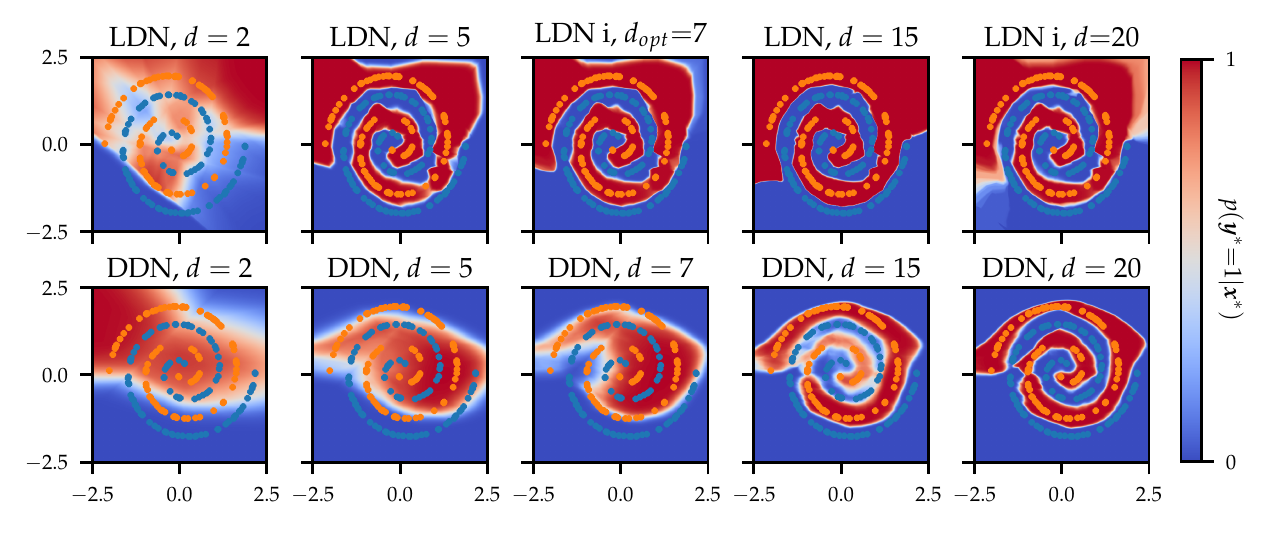}}
\vskip -0.05in
\caption{Top: spiral functions learnt at different depths of a LDN. The ``i'' indicator refers to the use of \cref{eq:predict_cutoff_depth} for predictions. Bottom: functions learnt at different depths of a DDN. In all cases $D\,{=}\,20$.}
\label{fig:input_space_DDN_LDN}
\end{center}
\vskip -0.2in
\end{figure}

We plot the functions learnt by different layers of a DDN in \cref{fig:input_space_DDN_LDN}. In excessively deep DDNs, intermediate layers contribute minimally. Only at layer 15 does the learnt function start to resemble a spiral. In LDNs, the first layers do most of the work. Layers after $d_{opt}$ learn functions close to the identity. This allows us to prune them, reducing computational cost at test time while obtaining the same test performance as when marginalising over all $D$ layers. This is shown in \cref{app:spirals}.

\subsection{Small Image Datasets} 

\begin{figure}[ht]
\begin{center}
\centerline{\includegraphics[width=\textwidth]{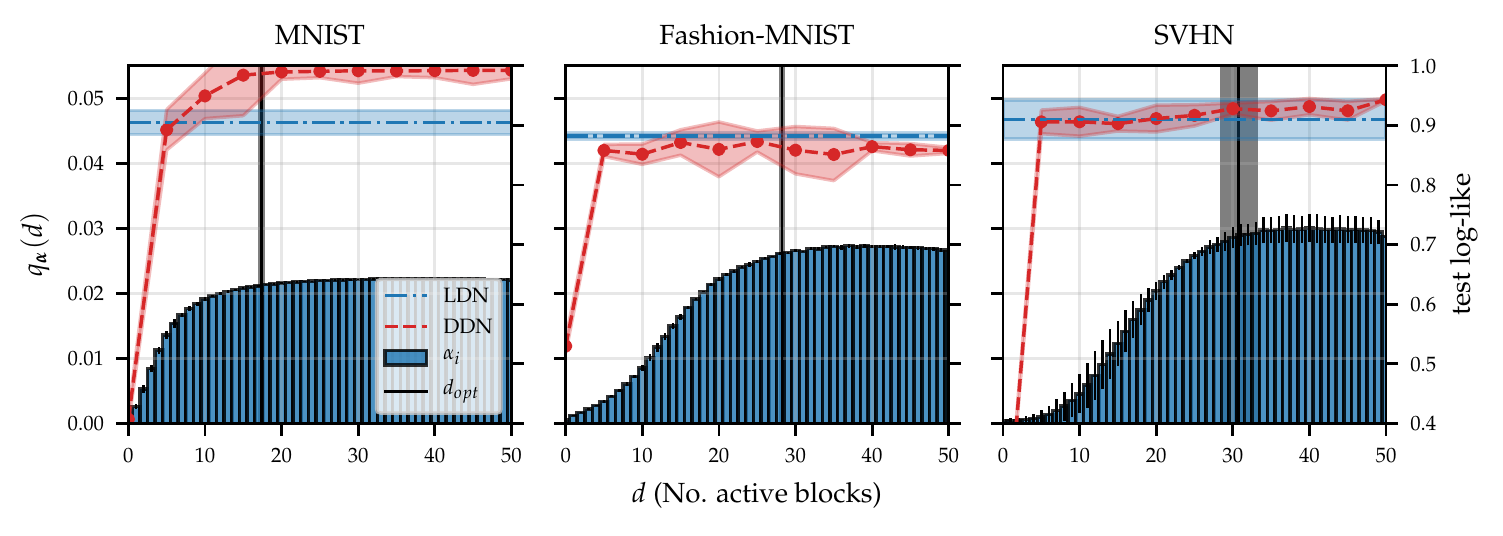}}
\vskip -0.1in
 \caption{Approximate posterior over depths for LDNs of $D\,{=}\,50$ trained on image datasets. Test log-likelihoods obtained for DDNs at various depths are overlaid with those from our LDNs when marginalising over the first $d_{opt}$ layers. $d_{opt}$ is chosen with the heuristic described in \cref{sec:proposed_method}.}
\label{fig:image_depth_dist}
\end{center}
\vskip -0.2in
\end{figure}

We further evaluate our approach on MNIST, Fashion-MNIST and SVHN. 
Each experiment is repeated 4 times. The results obtained with $D\,{=}\,50$ are shown in \cref{fig:image_depth_dist}. 
The larger size of these datasets diminishes the effect of the prior on the ELBO.
Models that explain the data well obtain large probabilities, regardless of their depth.
For MNIST, the probabilities assigned to each depth in our LDN grow quickly and flatten out around $d_{opt}\,{\approx}\,18$.
For Fashion-MNIST, depth probabilities grow slower. We obtain $d_{opt}\,{\approx}\,28$. For SVHN, probabilities flatten out around $d_{opt}\,{\approx}\,30$. These distributions and $d_{opt}$ values correlate with dataset complexity. 
In most cases, pruned LDNs achieve test log-likelihoods competitive with the best performing DDNs, while achieving equal or better accuracies, as shown in \cref{app:images}.
Additionally, our pruning strategy allows us to perform test-time inference approximately 62\%, 41\%, and 37\% faster than using $D\,{=}\,50$ layer networks on MNIST, Fashion-MNIST, and SVHN, respectively. We find pruning not to impact predictive performance.

We investigate the predictive uncertainty calibration of LDNs and DDNs on the datasets under consideration. Detailed results are found in \cref{app:images}. Similarly to \citet{network_callibration}, we find DDNs to be pathologically overconfident. LDNs present marginally better calibration on Fashion and SVHN, tending to be less overconfident for probabilities greater than 0.5. We observe a negligible degradation in calibration when pruning layers after $d_{opt}$. 

\section{Discussion and Future Work}\label{sec:discussion}

We formulate a variational objective over ResNet depth which can be evaluated exactly. It allows for one-shot learning of both model weights and a distribution over depth. We leverage this distribution to prune our networks, making test-time inference cheaper, and obtain model uncertainty estimates. Pruned networks obtain predictions competitive with regular networks of any depth on a toy spiral dataset, MNIST, Fashion-MNIST and SVHN. They also tend to provide better calibrated uncertainty estimates. Despite promising results, we have yet to evaluate the scalability of our approach to larger datasets. We leave this, and comparing to existing NAS approaches, to future work. We would also like to further investigate the uncertainty estimates given by depth-wise marginalisation.



  
  




\newpage
\bibliography{references}
\bibliographystyle{iclr2020_conference}

\newpage
\appendix
\section{Derivation of the ELBO in \cref{eq:var_objective} and Link to Variational EM}\label{app:derivation}

Referring to $\dset{=}\{\bff{X}, \bff{Y}\}$ with $\bff{X} = \{\bff{x}^{(n)}\}_{n=1}^{N} \text{, and } \bff{Y}=\{\bff{y}^{(n)}\}_{n=1}^{N}$ and, for simplicity, dropping sub-indices referring to functions' parameters $(\btheta, \balpha)$, we show that \cref{eq:var_objective} is a lower bound on $\log p(\dset{}) = \log p(\bff{Y} | \bff{X})$:
\begin{align}\label{eq:ELBO_derivation}
    \text{KL}(q(d)\,\|\,p(d|\dset)) &= \EX_{q(d)}[\log q(d) - \log p(d|\dset)] \notag \\
    &= \EX_{q(d)}\left[\log q(d) -  \log \frac{p(\bff{Y} | \bff{X}, d) \,  p(d)}{p(\bff{Y} | \bff{X})}\right]\notag \\
    &= \EX_{q(d)}[\log q(d) - \log p(\bff{Y} | \bff{X}, d) -\log p(d) + \log p(\bff{Y} | \bff{X})]\notag\\
    &= \EX_{q(d)}[-\log p(\bff{Y} | \bff{X}, d)] + \text{KL}(q(d)\,\|\,p(d)) + \log p(\bff{Y} | \bff{X})\notag\\
    &= -\mathcal{L}(\balpha, \btheta) + \log p(\bff{Y} | \bff{X}).
\end{align}
Using the non-negativity of the KL divergence, we can see that: $\mathcal{L}(\balpha, \btheta) \leq \log p(\bff{Y} | \bff{X})$.

We now show how our formulation corresponds to a variational EM algorithm~\citep{tzikas2008variational}. Here, network depth acts as the latent variable and network weights are parameters. For a given setting of network weights $\btheta^{k}$, at optimisation step $k$, we can obtain the exact posterior over $d$ using the \textit{E step}:
\begin{gather}\label{eq:e_step}
     \alpha^{k+1}_{j} = p(d{=}j|\dset, \btheta^{k}) = \frac{p(d=j)\cdot\prod_{n=1}^{N} p(\bff{y}^{(n)}|\bff{x}^{(n)}, d{=}j, \btheta^{k})}{\sum^{D}_{i=0}p(d{=}i)\cdot\prod_{n=1}^{N} p(\bff{y}^{(n)}|\bff{x}^{(n)}, d{=}i, \btheta^{k})}
\end{gather}
The posterior depth probabilities can now be used to marginalise this latent variable and perform maximum likelihood estimation of network parameters. This is the M step:
\begin{align}\label{eq:m_step}
     \btheta^{k+1} &=  \argmax_{\btheta} \EX_{p(d|\dset, \btheta^{k})}\left[\prod_{n=1}^{N} p(\bff{y}^{(n)}|\bff{x}^{(n)}, d, \btheta^{k})\right]\notag \\
     &= \argmax_{\btheta} \sum_{i=0}^{D} p(d{=}i|\dset, \btheta^{k}) \prod_{n=1}^{N} p(\bff{y}^{(n)}|\bff{x}^{(n)}, d{=}i, \btheta^{k})
\end{align}
Unfortunately, the E step \cref{eq:e_step} requires calculating the likelihood of the complete training dataset, an expensive operation when dealing with neural network models and big data. We sidestep this issue through the introduction of an approximate posterior $q(d)$, parametrised by ${\balpha}$, and a variational lower bound on the marginal log-likelihood \cref{eq:ELBO_derivation}. The corresponding variational E step is given by:
\begin{gather}\label{eq:var_E_step}
    \balpha^{k+1} = \argmax_{\balpha} \textstyle{\sum}_{n=1}^{N} \EX_{q_{\balpha}(d)}\left[ \log p(\bff{y}^{(n)}|\bff{x}^{(n)}, d, \btheta^{k})\right] - \text{KL}( q_{\balpha^{(k)}}(d)\,\|\,p_{\bbeta}(d))
\end{gather}
Because our variational family contains the exact posterior distribution - they are both categorical - the ELBO is tight at the optima with respect to the variational parameters $\balpha$. Solving \cref{eq:var_E_step} recovers $q_{\balpha^{k+1}}(d)\,{=}\,p(d|\dset, \btheta^{k})$. 
We now combine the variational E step \cref{eq:var_E_step} and M step \cref{eq:m_step}  updates, recovering \cref{eq:var_objective}, where $\balpha$ and $\btheta$ are updated simultaneously through gradient steps:
\begin{align*}
     \mathcal{L}(\balpha, \btheta) = \textstyle{\sum}_{n=1}^{N} \EX_{q_{\balpha}(d)}\left[ \log p(\bff{y}^{(n)}|\bff{x}^{(n)}, d, \btheta)\right] - \text{KL}( q_{\balpha}(d)\,\|\,p(d))
\end{align*}
This objective is amenable to minibatching. The variational posterior tracks the true posterior during gradient updates, providing an unbiased estimate. Thus, \cref{eq:var_objective}, allows us to optimise an unbiased estimate of the data's log-likelihood with network depth marginalised.

\section{Implementation Details}\label{app:implementation}

We implement all of our models in PyTorch~\citep{pytorch}. We train our models using SGD with a momentum value of $0.5$. The log-likelihood of the train data is obtained using the cross entropy function. We use the default PyTorch parameter initialisation in all experiments. We do not set specific random seeds. However, we run each experiment multiple times and obtain similar results, showing our approach's robustness to this parameter initialisation.

For our experiments on spirals, a fixed learning rate of $0.1$ and a batch size of $512$ are used. Note that for all experiments on spirals, except for the ones where the amount of training data is increased as part of the experiment, this results in full-batch gradient descent. Training progress is evaluated using the ELBO \cref{eq:var_objective}. Early stopping is applied after 500 iterations of not improving on the previous best ELBO. The parameter setting which provides the best ELBO is kept. We choose an exponentially decreasing prior to encourage shallower models:
\begin{align*}
 \beta_{i} = \frac{\gamma^{1 + i}}{\sum^{D}_{i=0} \gamma^{1 + i}}
\end{align*}
where $\gamma$ is set to $0.85$.

For MNIST~\cite{lecun2010mnist}, Fashion-MNIST~\citep{xiao2017/online}, and SVHN~\citep{netzer2011reading}, the same setup as above is used, with a few exceptions. Early stopping is applied after 30 iterations of not improving. Additionally, the learning rate is dropped from $0.1$ to $0.01$ after 30 iterations. Each data-set is normalised per-channel to have a mean of $0$ and a standard deviation of $1$. No other forms of data modification are used.

For our fully connected networks, used for spiral datasets, our input $f_{0}$ and output $f_{D+1}$ blocks consist of linear layers. These map from input space to the selected width $w$ and from $w$ to the output size respectively. Thus, selecting $d=0 \Rightarrow b_{i}{=}0\,\forall i \in [1, D]$ results in a linear model. The softmax operation is applied after $f_{D+1}$. The functions applied in residual blocks, $f_{i}(\cdot)\,\forall i \in [1, D]$, consist of a fully connected layer followed by a ReLU activation function and Batch Normalization~\citep{ioffe2015batch}.

Our CNN architecture uses a $5{\times}5$ kernel convolutional layer together with a $2{\times}2$ average pooling layer as an input block $f_{0}$. Due to the relatively small size of the images used in our experiments, no additional down-sampling layers are used in the convolutional blocks. The output block, $f_{D+1}$, is composed of a global average pooling layer followed by a fully connected residual block, as described in the previous paragraph, and a linear layer. The softmax operation is applied to output activations. The function applied in the residual blocks, $f_{i}(\cdot)\,\forall i \in [1, D]$, matches the preactivation bottleneck residual function described by \cite{preact_resnets} and uses $3{\times}3$ kernels. The outer number of channels is set to 64 and the bottleneck number is 32.

\section{Additional Experiments on 2D Spirals}\label{app:spirals}

\begin{figure}[ht]
\begin{center}
\centerline{\includegraphics[width=0.65\textwidth]{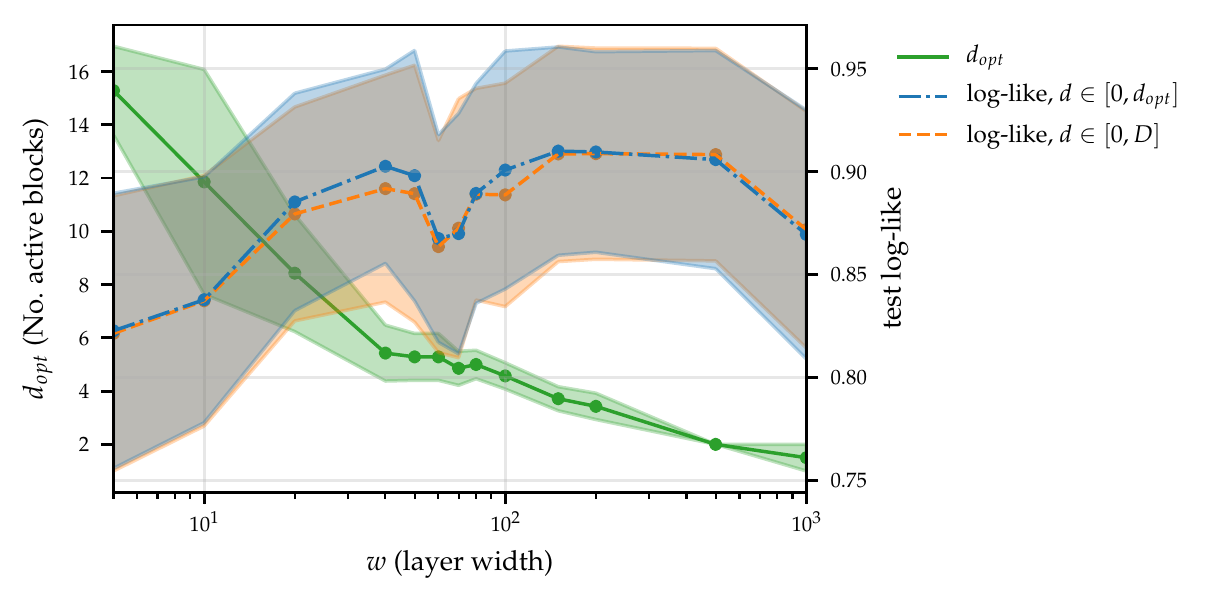}}
\vskip -0.1in
\caption[]{Evolution of LDNs' chosen depth and test performance as their layer width $w$ increases. The results obtained when making predictions by marginalising over all $D\,{=}\,20$ layers overlap with those obtained when only using the first $d_{opt}$ layers. The x-axis is presented in logarithmic scale.}
\label{fig:spiral_width_increase}
\end{center}
\vskip -0.15in
\end{figure}

We further explore the properties of LDNs in the small data regime by varying the layer width $w$. As shown in \cref{fig:spiral_width_increase}, very small widths result in very deep LDNs and worse test performance. Increasing layer width gives our models more representation capacity at each layer, causing the learnt depth to decrease rapidly. Test performance remains stable for widths in the range of $20$ to $500$, showing that our approach adapts well to changes in this parameter. The test log-likelihood starts to decrease for widths of $1000$, possibly due to training instabilities.

\begin{figure}[ht]
\begin{center}
\centerline{\includegraphics[width=0.75\textwidth]{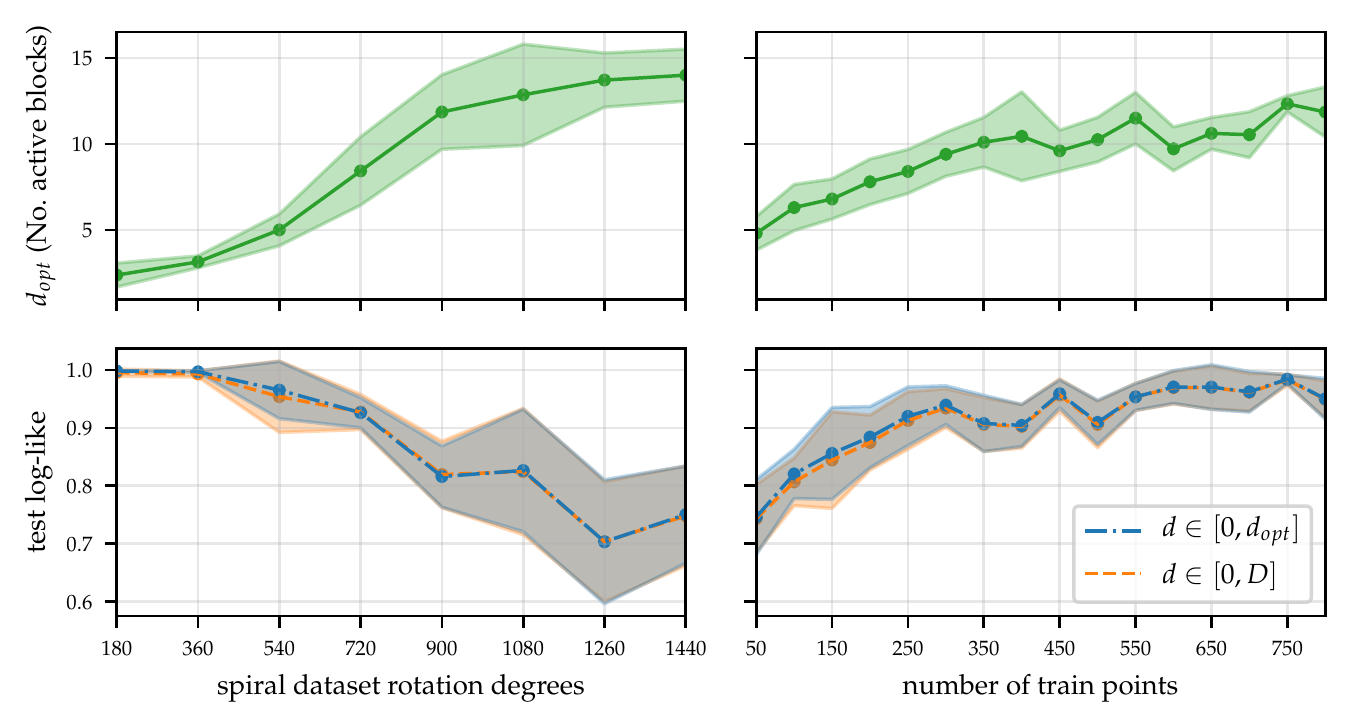}}
\vskip -0.1in
\caption[]{The left-side plots show the evolution of test performance and learnt depth as the data complexity increases. The right side plots show changes in the same variables as the number of train points increases. The results obtained when making predictions by marginalising over all $D\,{=}\,20$ layers overlap with those obtained when only using the first $d_{opt}$ layers. Best viewed in colour.}
\label{fig:rotation_Npoints}
\end{center}
\vskip -0.15in
\end{figure}

\begin{figure}[ht]
\begin{center}
\centerline{\includegraphics[width=0.9\textwidth]{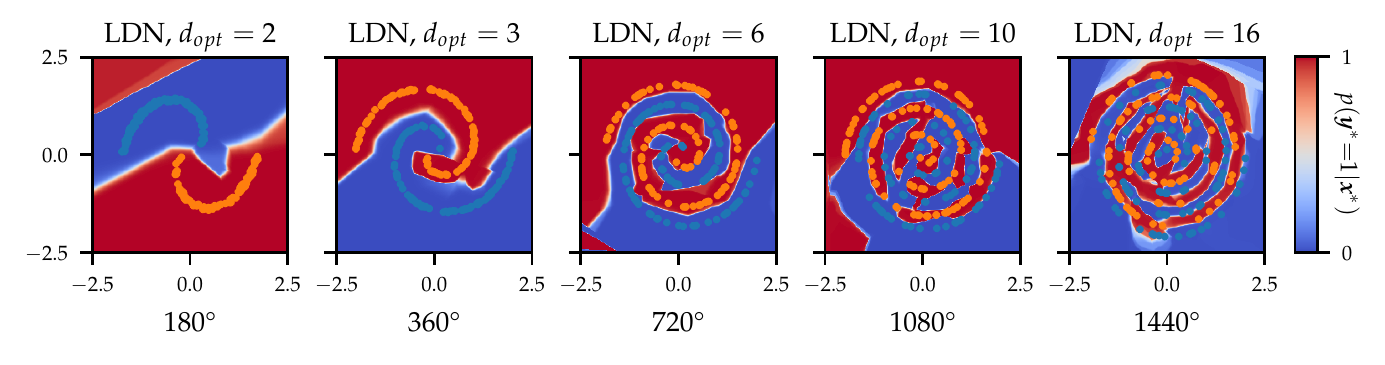}}
\vskip -0.1in
\caption[]{Functions learnt at each depth of a LDN on increasingly complex spirals. Note that single depth settings are being evaluated in this plot. We are not marginalising all layers up to $d_{opt}$.}
\label{fig:spiral_data_complex_input_scan}
\end{center}
\vskip -0.2in
\end{figure}

Setting $w$ back to $20$, we generate spiral datasets (code given in repo) with varying degrees of rotation while keeping the number of train points fixed to $200$. In \cref{fig:rotation_Npoints}, we see how LDNs increase their depth to match the increasing complexity of the underlying generative process of the data. For rotations larger than 720\degree, $w\,{=}\,20$ may be excessively restrictive. Test performance starts to suffer significantly. \cref{fig:spiral_data_complex_input_scan} shows how our LDNs struggle to fit these datasets well.

\begin{figure}[ht]
\begin{center}
\centerline{\includegraphics[width=0.9\textwidth]{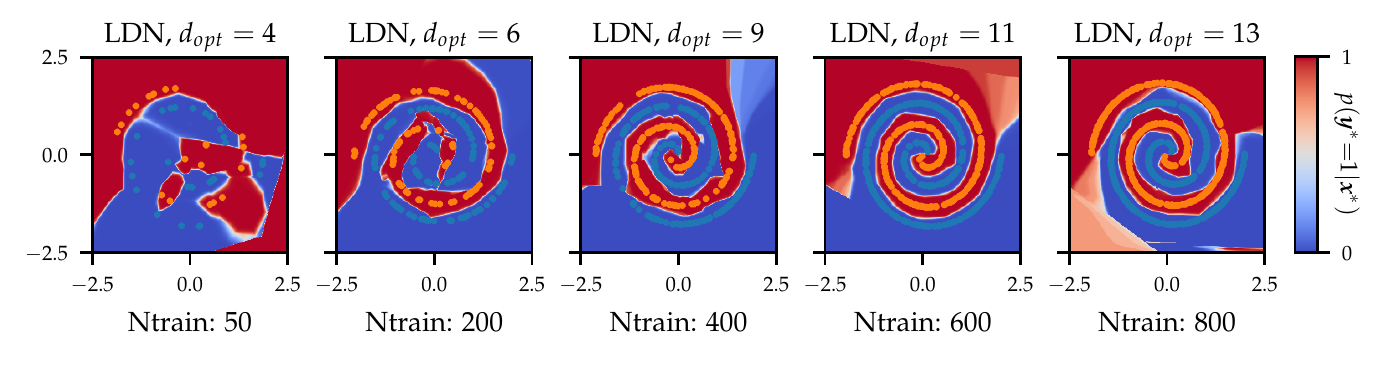}}
\vskip -0.1in
\caption[]{Functions learnt by LDNs trained on increasingly large spiral datasets. Note that single depth settings are being evaluated in this plot. We are not marginalising all layers up to $d_{opt}$.}
\label{fig:spiral_Ndata_input_scan}
\end{center}
\vskip -0.2in
\end{figure}

Returning to 720\degree spirals, we vary the number of training points in our dataset. We plot the LDNs' learnt functions in \cref{fig:spiral_Ndata_input_scan}. LDNs overfit the 50 point train set but, as displayed in \cref{fig:spiral_data_complex_input_scan}, learn very shallow network configurations. Increasingly large training sets allow the LDNs to become deeper while increasing test performance. Around 500 train points seem to be enough for our models to fully capture the generative process of the data. After this point $d_{opt}$ oscillates around 11 layers and the test performance remains constant. Marginalising over $D$ layers consistently produces the same test performance as only considering the first $d_{opt}$. All figures are best viewed in colour.

\clearpage

\section{Additional Experiments on Image Datasets}\label{app:images}

\begin{figure}[ht]
\vskip -0.2in
\begin{center}
\centerline{\includegraphics[width=0.9\textwidth]{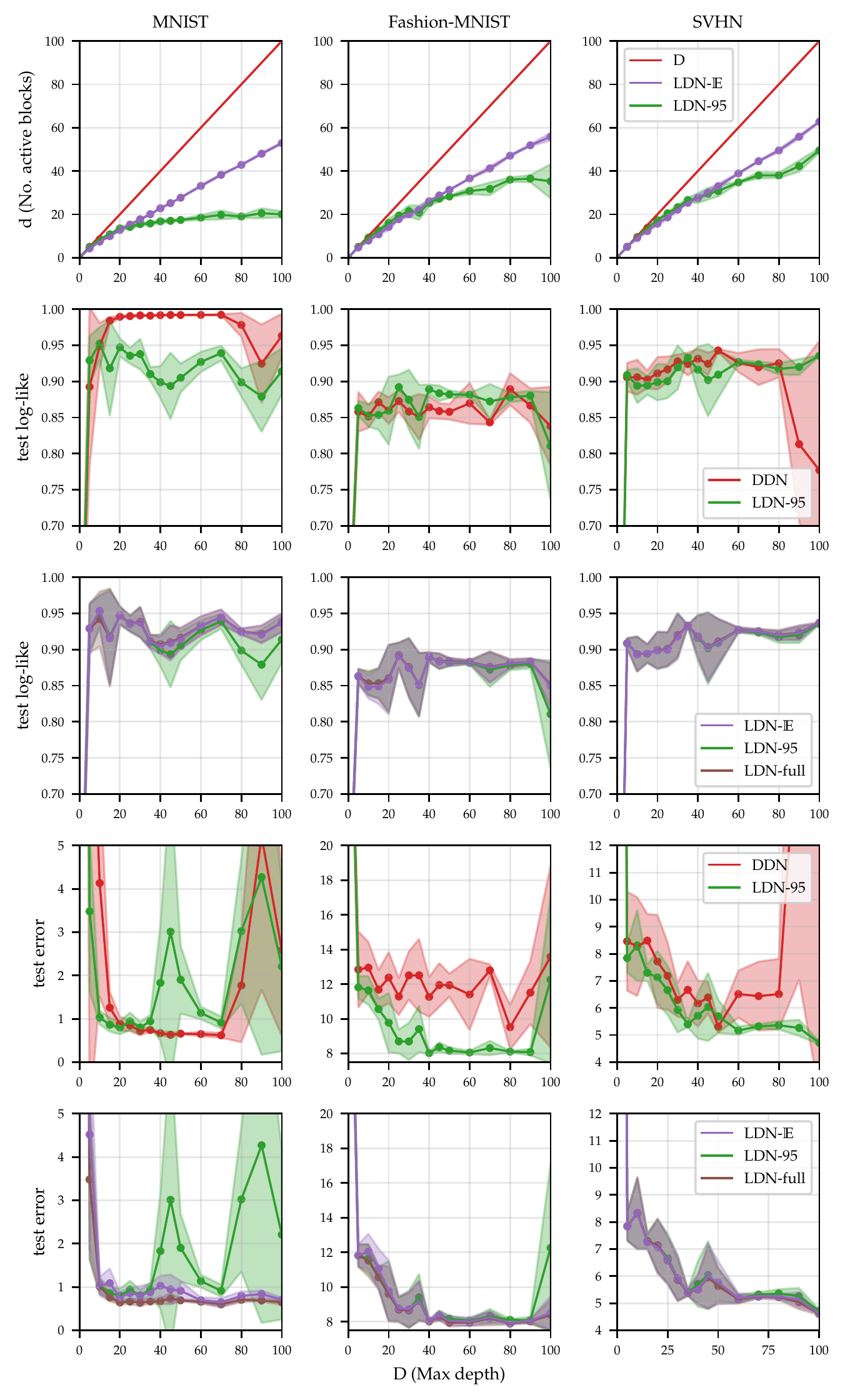}}
\vskip -0.1in
 \caption{Comparisons of DDNs and LDNs using different pruning strategies and maximum depths. \emph{LDN-95} refers to the pruning strategy described in \cref{sec:proposed_method}. \emph{LDN-$\mathbb{E}$} refers to pruning to the expected depth given by $\text{round}(\mathbb{E}_{q_{\balpha}}[d])$. \emph{LDN-full} refers to an unpruned LDN. \nth{1} row: comparison of learnt depth. \nth{2} row: comparison of test log-likelihoods for DDNs and LDNs with \nth{95} percent pruning. \nth{3} row: comparison of test log-likelihoods for LDN pruning methods. \nth{4} and \nth{5} rows: as above but for test error. Best viewed in colour.}
\label{fig:image_depth_comparison}
\end{center}
\vskip -0.15in
\end{figure}

\Cref{fig:image_depth_comparison} shows more detailed experiments comparing LDNs with DDNs on image datasets. We introduce expected depth $d_{opt}\,{=}\,\text{round}(\EX_{q_{\balpha}}[d])$ as an alternative to the \nth{95} percent heuristic introduced in \cref{sec:proposed_method}. The first row of the figure adds further evidence that the depth learnt by LDNs corresponds to dataset complexity. For any maximum depth, and both pruning approaches, the LDN's learnt depth is smaller for MNIST than Fashion-MNIST and likewise smaller for Fashion-MNIST than SVHN. For MNIST, Fashion-MNIST and, to a lesser extent, SVHN the depth given by the \nth{95} percent pruning tends to saturate. On the other hand, the expected depth grows with $D$, making it a less suitable pruning strategy. 

As shown in rows 2 to 5, for SVHN and Fashion-MNIST, \nth{95} percentile-pruned LDNs suffer no loss in predictive performance compared to expected depth-pruned and even non-pruned LDNs. They outperform DDNs. For MNIST, \nth{95} percent pruning gives results with high variance and reduced predictive performance in some cases. Here, DDNs yield better log-likelihood and accuracy results. Expected depth is more resilient in this case, matching full-depth LDNs and DDNs in terms of accuracy.

\begin{figure}[ht]
\begin{center}
\begin{subfigure}[b]{\textwidth}
    \centerline{\includegraphics[width=0.85\textwidth]{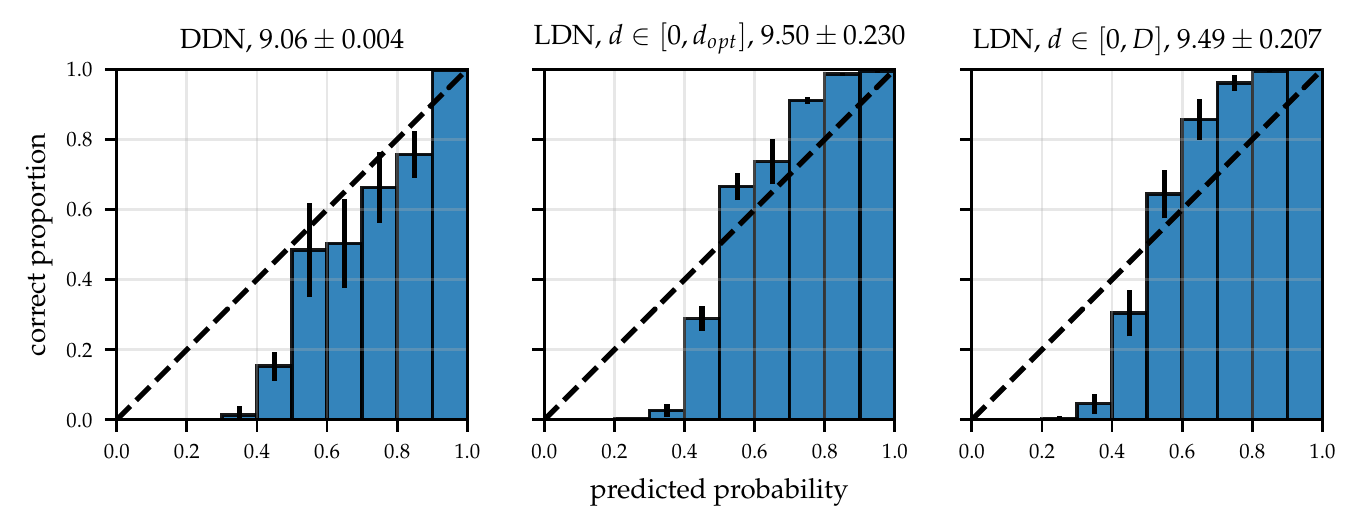}}
    \caption{MNIST, $d_{opt}\,{\approx}\,18$}
    \label{fig:mnist_callibration}
\end{subfigure}

\begin{subfigure}[b]{\textwidth}
    \centerline{\includegraphics[width=0.85\textwidth]{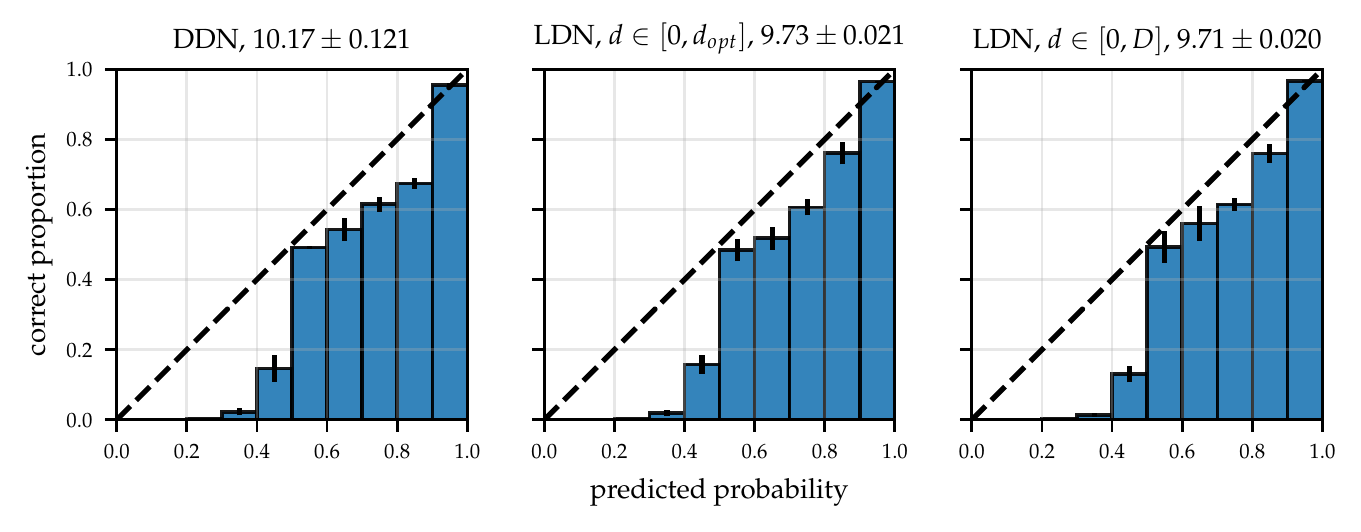}}
    \caption{Fashion-MNIST, $d_{opt}\,{\approx}\,28$}
    \label{fig:fmnist_callibration}
\end{subfigure}

\begin{subfigure}[b]{\textwidth}
    \centerline{\includegraphics[width=0.85\textwidth]{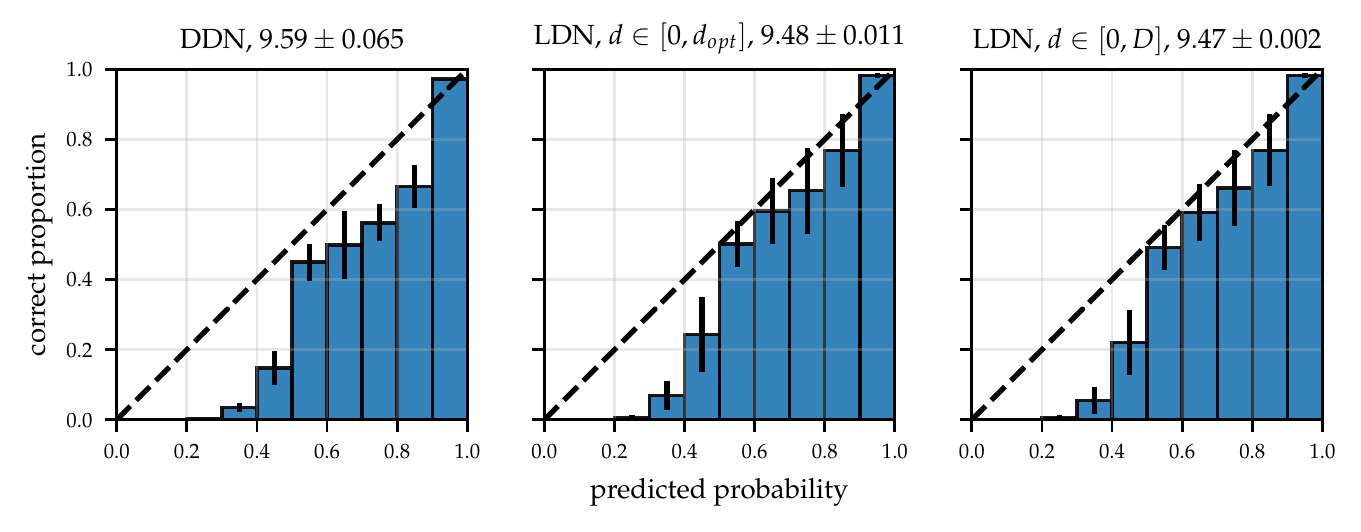}}
    \caption{SVHN, $d_{opt}\,{\approx}\,30$}
    \label{fig:svhn_callibration}
\end{subfigure}
\end{center}
\vskip -0.2in
\caption{Calibration plots obtained for image datasets. Results for a 50 layer DDN, $d_{opt}$ layer LDN and $50$ layer LDN, are shown on the left, centre, and right respectively. The expected calibration errors corresponding to each plot are given in the titles. All models have a max depth of $D\,{=}\,50$.}
    \label{fig:image_callibration}
\end{figure}

\cref{fig:image_callibration} shows calibration results for the image datasets under consideration. In all cases, DDNs are overconfident for all predicted probabilities. For Fashion-MNIST and SVHN, LDNs present less overconfidence for probabilities greater than $0.5$. They achieve lower expected calibration errors \citep{network_callibration} overall. On MNIST, LDNs present strong underconfidence for probabilities larger than $0.5$. Their calibration error is worse than that of DDNs. Together with the results from \cref{fig:image_depth_comparison}, this suggests that our LDNs are underfitting on MNIST. In all cases, the expected calibration errors of pruned LDNs are marginally larger than those of non-pruned LDNs. 

\begin{figure}[ht]
\begin{center}
\centerline{\includegraphics[width=0.85\textwidth]{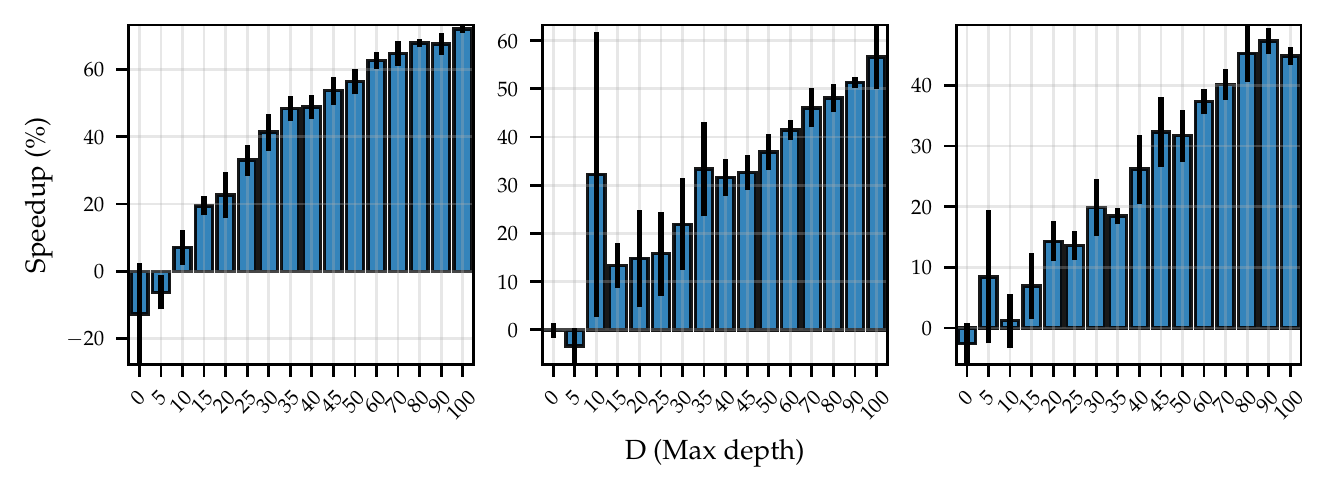}}
\vskip -0.1in
 \caption{Proportional speedup for a single forward pass obtained with $d_{opt}$ layer LDNs compared to their $D$ layer DDN counterparts.}
\label{fig:image_speedups}
\end{center}
\vskip -0.15in
\end{figure}

\cref{fig:image_speedups} shows the proportional reduction in forward pass time for pruned LDNs relative to DDNs, both having the same maximum depth $D$. In line with our expectations, the speedup provided by our proposed \nth{95} percent pruning strategy grows with $D$. For $D\,{=}\,50$, we obtain  62\%, 41\%, and 37\% speedups for MNIST, Fashion-MNIST, and SVHN respectively.

\clearpage

\section{The NAS Best Practices Checklist}\label{app:checklist}

\subsection{Best practices for releasing code}

\begin{itemize}
    \item[\Checkedbox] Code for the training pipeline used to evaluate the final architectures
    \item[\Checkedbox] Code for the search space
    \item[\Checkedbox] The hyperparameters used for the final evaluation pipeline, as well as random seeds
     \begin{itemize}
        \item Our evaluation pipeline has no random seeds or hyperparameters.
    \end{itemize}
    \item[\Checkedbox] Code for your NAS method
    \item[\HollowBox] Hyperparameters for your NAS method, as well as random seeds
    \begin{itemize}
        \item We report all hyperparameters but not random seeds. We run our methods multiple times in all of our experiments and obtain similar results.
    \end{itemize}
\end{itemize}

\subsection{Best practices for comparing NAS methods}

\begin{itemize}
    \item[\Checkedbox] For all NAS methods you compare, did you use exactly the same NAS benchmark,  including  the  same dataset (with  the  same  training-test  split), search space and code for  training  the  architectures  and hyperparameters for that code?
    \item[\Checkedbox] Did you control for confounding factors (different hardware, versions of DL libraries, different run times for the different methods)?
    \item[\CrossedBox]  Did you run ablation studies?
    \begin{itemize}
        \item Not applicable for our method.
    \end{itemize}
    \item[\Checkedbox] Did you use the same evaluation protocol for the methods being compared?
    \item[\CrossedBox] Did you compare performance over time?
    \begin{itemize}
        \item Not applicable for our method and benchmarks.
    \end{itemize}
    \item[\CrossedBox] Did you compare to random search?
    \begin{itemize}
        \item Our search space is one dimensional allowing us to perform grid search.
    \end{itemize}
    \item[\HollowBox] Did you perform multiple runs of your experiments and report seeds?
    \begin{itemize}
        \item While we did perform multiple runs of our experiments, we did not report the seeds.
    \end{itemize}
    \item[\CrossedBox] Did you use tabular or surrogate benchmarks for in-depth evaluations?
    \begin{itemize}
        \item Not applicable for our search space.
    \end{itemize}
\end{itemize}

\subsection{Best practices for reporting important details}

\begin{itemize}
    \item[\CrossedBox] Did you report how you tuned hyperparameters, and what time and resources this required?
    \begin{itemize}
        \item Our approach has the same non-architecture hyperparameters as training a regular neural network: learning rate, learning rate decay, early stopping, and batch size. We used standard values as reported in \cref{app:implementation}. We did not perform hyperparameter tuning. For our spiral experiments, we ran an experiment comparing performance across network widths, see \cref{app:spirals}.  
    \end{itemize}
    \item[\CrossedBox] Did you report the time for the entire end-to-end NAS method (rather than, e.g., only for the search phase)?
    \begin{itemize}
        \item Both times are equivalent in our case as it is a one-shot method. Our approach takes the same time as training a regular ResNet. 
    \end{itemize}
    \item[\Checkedbox] Did you report all the details of your experimental setup?
\end{itemize}

\end{document}